\documentclass[letterpaper, 10 pt, conference]{ieeeconf}  
\usepackage{algpseudocode}
\usepackage{algorithm}
\usepackage{amsmath,amsfonts,amssymb}
\usepackage{bm}
\usepackage{cite}
\usepackage{lipsum}
\usepackage{graphicx}
\usepackage[font=small,skip=10pt]{caption}
\usepackage{subcaption}
\usepackage{tikz}
\usepackage{pgfplots}
\usepackage{csvsimple}
\usepackage{filecontents}
\usepackage{balance}
\usepackage[colorlinks=true]{hyperref}
\hypersetup{urlcolor=blue, linkcolor=blue}

\IEEEoverridecommandlockouts                              

\title{\LARGE \bf
On the Collaborative Object Transportation Using Leader Follower Approach
}

\author{Sumanta Ghosh$^{1*}$, Subhajit Nath$^{1*}$, Sarvesh Sortee$^{1}$, Lokesh Kumar$^{1}$, and Titas Bera$^{1}$ 
\thanks{* Sumanta Ghosh and Subhajit Nath contributed equally to this work.}
\thanks{$^{1}$Sumanta Ghosh, Subhajit Nath, Sarvesh Sortee, Lokesh Kumar, Titas Bera with Robotics and Autonomous Systems Group, TCS Research, India
        {\tt\small \{g.sumanta1,nath.subhajit, sarvesh.sortee, kumar.lokesh7, titas.bera\}@tcs.com}}%
}

\begin{document}

\maketitle
\thispagestyle{empty}
\pagestyle{empty}

\begin{abstract}

In this paper we address the multi-agent collaborative object transportation problem in a partially known environment with obstacles under a specified goal condition. We propose a leader follower approach for two mobile manipulators collaboratively transporting an object along specified desired trajectories. The proposed approach treats the mobile manipulation system as two independent subsystems: a mobile platform and a manipulator arm and uses their kinematics model for trajectory tracking. In this work we considered that the mobile platform is subject to non-holonomic constraints, with a manipulator carrying a rigid load. The desired trajectories of the end points of the load are obtained from Probabilistic RoadMap-based planning approach. Our method combines Proportional Navigation Guidance-based approach with a proposed Stop-and-Sync algorithm to reach sufficiently close to the desired trajectory, the deviation due to the non-holonomic constraints is compensated by the manipulator arm. A leader follower approach for computing inverse kinematics solution for the position of the end-effector of the manipulator arm is proposed to maintain the load rigidity. Further, we compare the proposed approach with other approaches to analyse the efficacy of our algorithm.

\end{abstract}

\vspace{3mm}
\section{INTRODUCTION}

With the rapid pace of recent developments in robotics, its applications have become extensive. Given the general advantages of cooperative mobile manipulators over a single mobile manipulator; vast studies have been carried out in recent years over the abilities of the same, mainly to accomplish complex tasks like object transportation, assembling, packaging and welding\cite{feng2020overview,khatib1996coordination,song2002potential}. Most of the tasks mentioned above are repetitive in nature and performed in structured environment. Centralized control approaches are very popular for solving this kind of problem though they suffered from high computational complexities \cite{alford1984coordinated,rus1995moving,bohringer1997distributed}. Decentralized and distributed control strategies are proposed to mitigate this problem \cite{donald2000distributed,marino2017distributed,habibi2015distributed,zhu2020trajectory}. A decentralised approach to multi-robot manipulation in which a team of robots encircles and traps an object and transports it by dragging and pushing to the target configuration in an obstacle laden environment is presented in \cite{fink2008multi}. A formalism of local motion planning for collaborative multi-robot manipulation was developed in \cite{alonso2015local}. It exploits the deformability of soft objects during manipulation. Formation based control approaches are very popular in collaborative multi-agent load transportation.
\begin{figure}[t]
    \centering
    \includegraphics[width=1.\linewidth, height=0.8\linewidth]{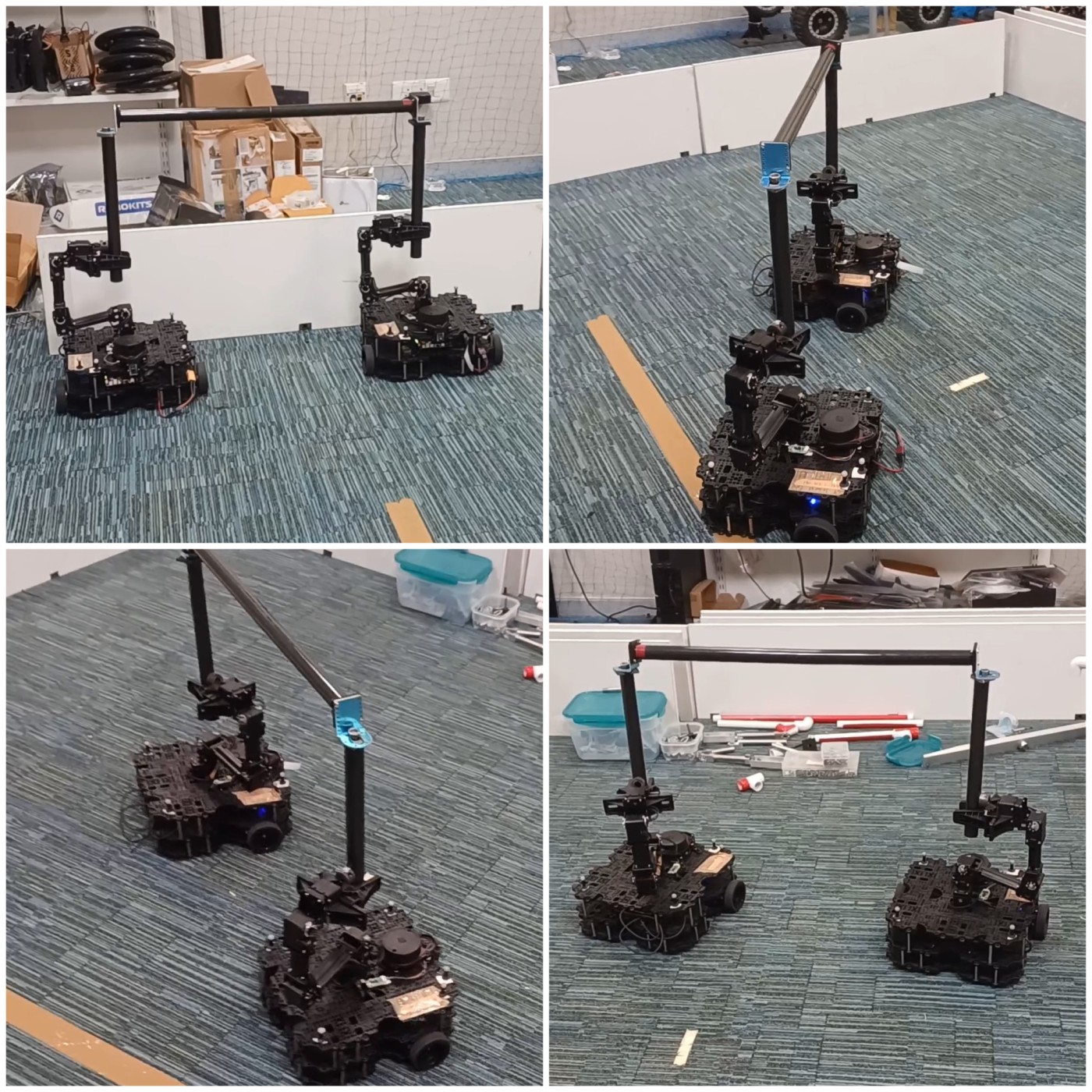}
    \caption{Collaborative transportation of a rigid rod by two mobile manipulators in a circular trajectory.}
    \label{f1}
    \vspace{-1.5em}
\end{figure}
In \cite{alonso2015multi}, a sequential convex programming formulation is proposed to solve a formation control problem for collaborative transportation of a rigid object by two mobile manipulators in dynamic environment. Authors in \cite{abbaspour2015optimal} presented a method for an optimal formation of a team of non-holonomic robots for a common object manipulation. A distributed swarm algorithm with application in multi-agent collaborative load transportation is developed in \cite{carpio2018distributed} where agents are reaching an aggregate state maintaining the environment constraints.      Authors in \cite{bujarbaruah2021learning} proposed a decentralized leader-follower  hierarchical strategy for two robots collaboratively transporting an object in a partially known environment with obstacles.

In this work we consider the problem of object transportation using mobile manipulators where two robots are transporting a rigid rod collaboratively complying with object's physical constraints in an environment with obstacles. The desired trajectories of the end points of the object are generated using a global sampling based path planner. 
We consider that the measure of success while attempting to reach the desired goal location is not the the time taken or the trajectory error, rather how close the distance between the end effectors of the robots remains to the length of the load during the execution, while staying as close to the desired trajectory as possible. We propose a Stop-and-Sync algorithm, combined with Proportional Navigation Guidance (PNG) based approach to steer the mobile manipulator close to desired the trajectory, and a leader follower approach for solving inverse kinematics to track the desired trajectory. 

The main contribution of this paper is developing an efficient method for a dual robot system to transport a rigid object from a specified initial position to a final position, while avoiding collisions with obstacles maintaining the synchronization between the agents.


The remainder of the paper is organized as follows, Section \ref{sec2} introduces the problem formulation and a overview of the method. In Section \ref{sec3} we introduce a Stop-and-Sync algorithm along with the PN Guidance based method. The proposed leader follower approach is discussed at the later part of that section. Section \ref{sec4} contains the simulation results. We conclude the paper with discussion in Section \ref{sec5}.   

\section{Problem Formulation \& Approach}
\label{sec2}
We represent the position of a point $B$ measured from a point $A$, expressed in frame $F$ as $^A\textbf{\textit{p}}_F^B$. When the positions are expressed in the world frame, for the ease of notation, we denote the position of point $B$, measured from $A$, as $^A\textbf{\textit{p}}^B$. Further, to represent the position vector of a point $B$ measured from the origin of the world frame expressed in the world frame we use $\textbf{\textit{p}}^B$.

\begin{figure}
\centering
	{\includegraphics[scale=0.5, keepaspectratio = true]{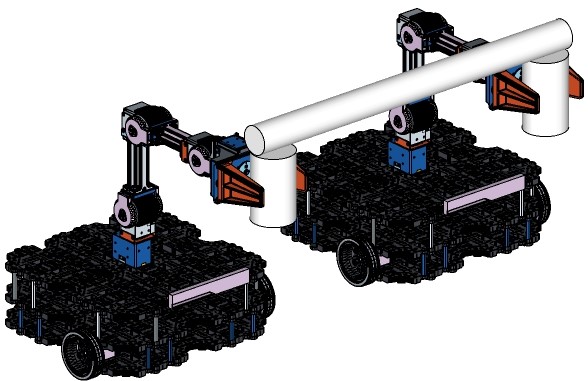}}\hspace{0.1cm}
    \caption{ \small Schematic Representation of load transport with two TurtleBot3 Waffle Pi mounted with the OpenMANIPULATOR-X arm. The specifications for the same can be found \href{https://emanual.robotis.com/docs/en/platform/turtlebot3/features/}{here}. 
    Due to orientation constraints of the end-effector, the rigid load is supported by two vertical fixtures connected by revolute joints.}
    \label{turtlebot_fig}
\end{figure}
We consider a system of two mobile manipulators carrying a rigid cylindrical load connected through their end-effector, as shown in Fig.(\ref{turtlebot_fig}). The aim is to navigate the load from an initial position to a final while avoiding obstacles and keeping the distance between the end-effectors same as the length of the rigid load for all the time. 

The robot which acts as the Leader is denoted by subscript $L$, while the Follower is expressed by the subscript $F$. The desired end-effector trajectory to be followed by each $a^{\text{th}}$ robot is denoted as: $T_a$; whose projection on the $X-Y$ plane is denoted as $P_a$. We assume that the desired end-effector trajectory for each robot has $n_d$ number of way-points. These way-points, when projection onto the $X-Y$ plane, act as target points for the corresponding mobile bases. Accordingly, these projected desired target points are represented as: $\textbf{d}^a_P = (x^{d,a}_P, y^{d,a}_P)$. The desired end-effector positions at each time step are obtained by projection to find the nearest point on $T_a$. We discuss more about it later in Section \ref{sec3}. These positions are represented as: $\textbf{d}^a_n = (x^{T,a}_n, y^{T,a}_n, z^{T,a}_n)$. The base locations at each time step $k$ are given as: $\textbf{b}^a_k = (x^a_k, y^a_k)$.

We assume that $T_a$ is generated from PRM-based path planner considering the obstacles. We consider that the base of the mobile manipulators are non-holonomic and the motion of the mobile platform is confined within XY plane only. The mobile manipulator system consists of two main  subsystems: the platform or base of the mobile manipulator, and the manipulator arm. The goal is to start the $a^{\text{th}}$ robot from a position $(x_k^a,y_k^a)$, sufficiently close to $\textbf{d}^a_P$, compute the control inputs, $(v^a,\omega^a)$ to the mobile base that lands the mobile manipulator to a point $(x^a_{k+1},y^a_{k+1})$ within the reachability distance of the next point $(x^d_{k+1},y^d_{k+1})$ on the desired trajectory. The reachability distance of a point is defined as follows:
\begin{equation}
    \label{pd}
    \rho_d=\gamma*\rho_l
\end{equation}
Where, $\rho_l$ is the reachability of a manipulator defined as the volume of the workspace that can be reached by the manipulator. $\gamma$ is a user defined value that represents a tolerable fraction of the reachability. Henceforth, the joint angles of the manipulator arm are calculated by solving the inverse kinematics of the manipulator arm to the desired point $\textbf{d}^a_n$ measured from the base of the mobile manipulator expressed in the world frame depending on the role (leader or follower) of the manipulator. The whole process is performed for every base position $\textbf{b}^a_k$ to track the desired end-effector trajectory for a single robot. Stop-and-Sync algorithm is introduced to maintain the collaboration in between them.   

\subsection{Path Planning}
We consider that a 3D model of the environment with polyhedron obstacles is present. The position, orientation, and the shape of all the static obstacles are assumed to be known apriori.
\begin{figure}
\centering
\subfloat[]{\includegraphics[width=0.5\linewidth,height=3cm,keepaspectratio=true]{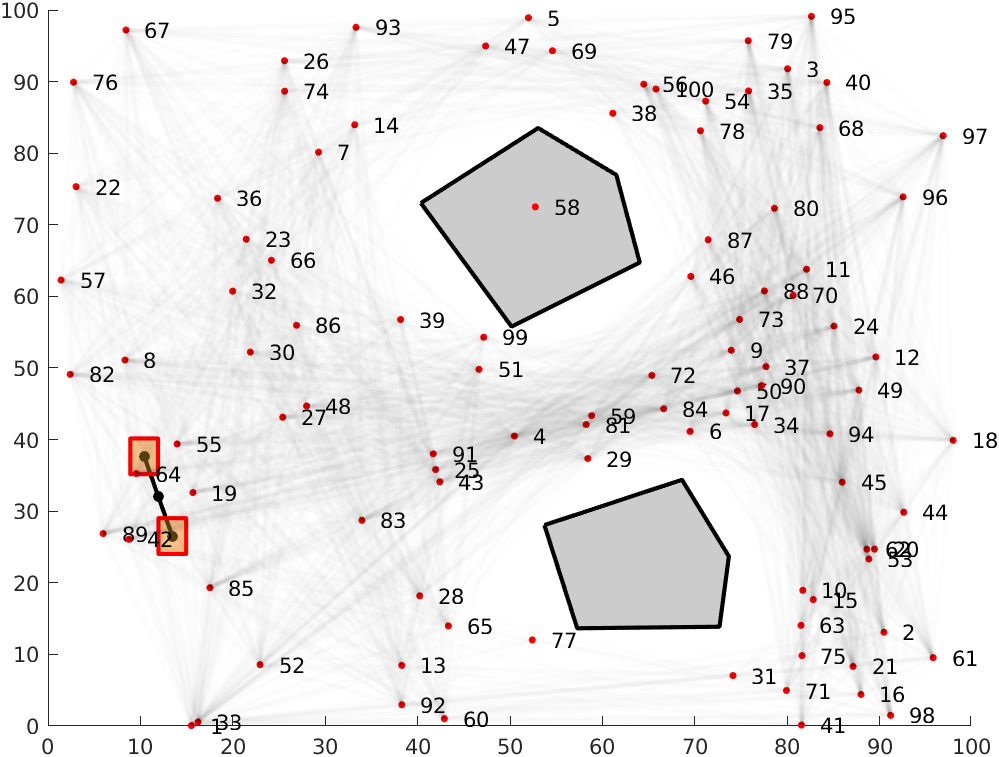}}\hspace{0.25cm}
\subfloat[]{\includegraphics[width=0.5\linewidth,height=3cm,keepaspectratio=true]{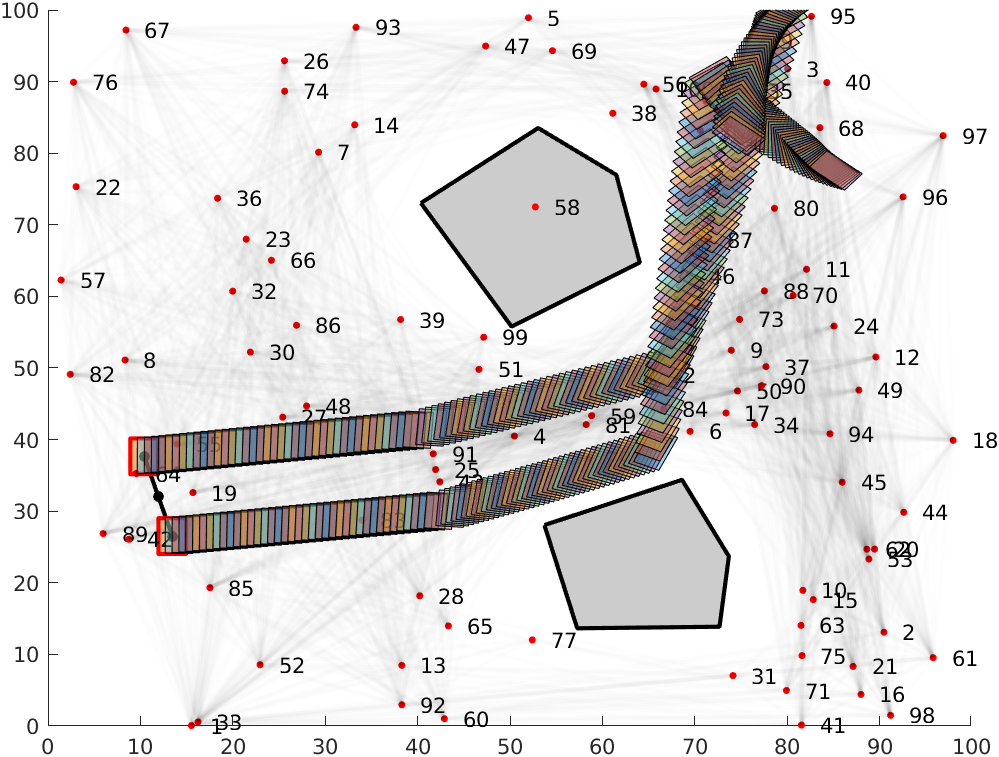}}
\caption{\small Schematic of the nodes generated through PRM in (a), and the path planning achieved in (b). The grayed areas represent the obstacles, while the robots carrying the load are represented by the red squares.}
\label{PRM_plan}
\end{figure}

For path planning, we use PRM on a 2D projection of the above stated obstacle laden environment. To that end, given the initial and final configurations $q_{init}$ and $q_{goal}$, a number of vertices are generated randomly over $Q_{free}$ -- which are then connected to its $c$ closest neighbours via edges using a local planner. Henceforth, the shortest path between $q_{init}$ and $q_{goal}$ through the obtained edges is extrapolated using Dijkstra’s Algorithm.
The execution of this algorithm in MATLAB is shown in Fig. \ref{PRM_plan}.  

\subsection{Forward Kinematics of Mobile Platform}
For a mobile robot the forward kinematics is defined as the mapping between the control variables (as a function of time, $t$) and the pose of the robot. The approach to obtain a mapping between $\textbf{b}^a_k$ and the control variables is the same as followed in []. In our case, the control variables are the linear and the angular velocities of the mobile robot. Assuming the trajectory covered by the mobile base from timesteps $k$ to $k+1$ to be composed of small circular segments; for a differential drive robot starting with a pose $[x(t), y(t), \theta(t)]^T$, moving with linear velocity $v(t)$ and angular velocity $\omega(t)$, the pose of the robot at $(t+\delta t)$ can be determined from the forward kinematics,

\begin{equation}
    \label{FK}
    \begin{bmatrix}
    x(t+\delta t)\\
    y(t+\delta t)\\
    \theta(t+\delta t)
    \end{bmatrix}=\textbf{\textit{g}}^B(v(t),\omega(t))
\end{equation}
\begin{figure*}[t]
    \centering
    \includegraphics[width=7 in,height=9cm]{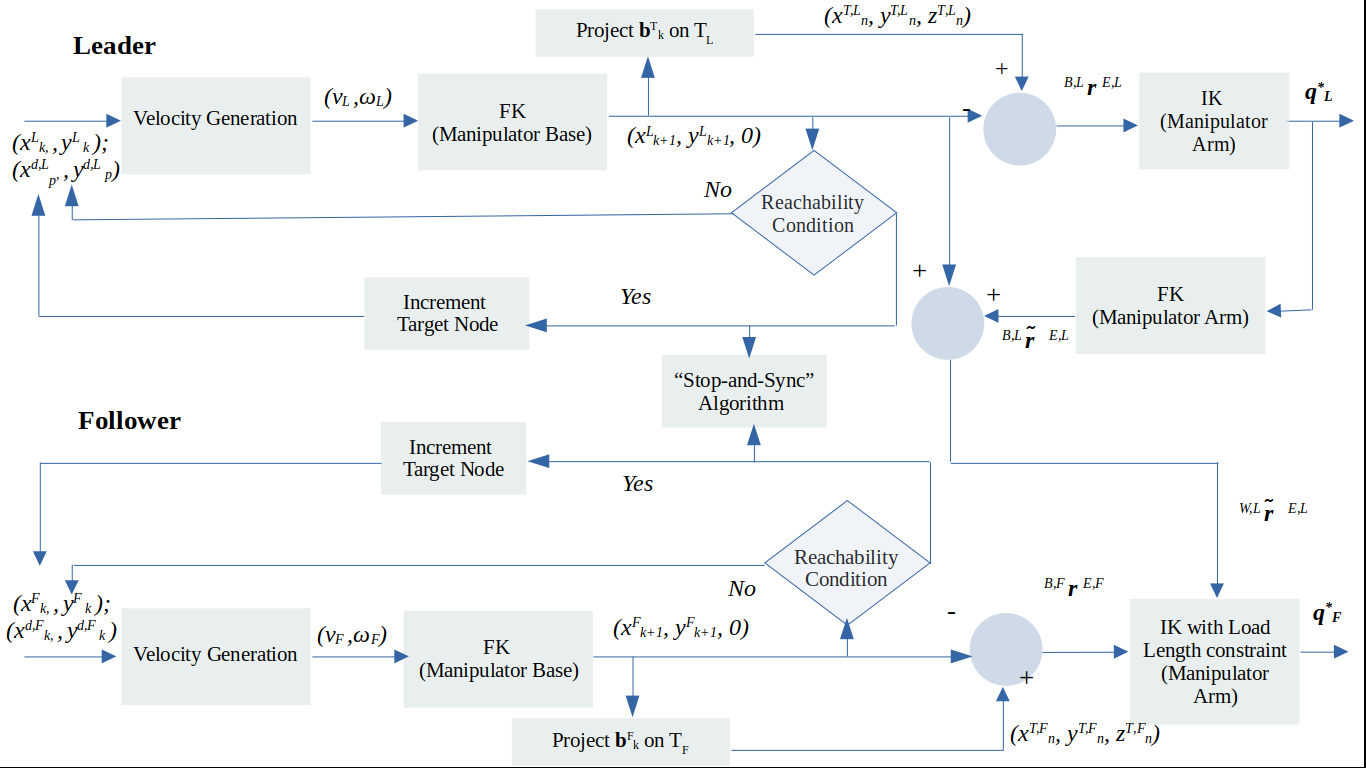}
    \caption{Schematic diagram of the Leader Follower Approach with Stop-and-Sync Algorithm. }
    \label{lf}
\end{figure*}

where the matrix $\textbf{g}^B$ was as derived in \cite{chong1997accurate}. Appropriate limits can be introduced to replace the circular trajectory with a straight line path. 

\subsection{Inverse Kinematics of Manipulator Arm}

The forward kinematics of a manipulator provides the position and the orientation of the end effector given the joint variables in our case the joint angles, $\textbf{\textit{r}}^E = \textbf{\textit{f}}^E\left(\bm{\beta}\right)$. This mapping from the joint space to the end effector position is highly nonlinear in nature. The problem of solving the inverse mapping, from the end effector position to the joint angles, is known as inverse kinematics of the manipulator, $\bm{\beta}=\textbf{\textit{f}}^{-1}\left(\textbf{\textit{r}}^E\right)$. Various analytic and numerical approaches exist for finding the solution of the inverse kinematics problem. One efficient approach is to formulate the problem as an optimization problem,
\begin{align} 
    \label{IK}
    &\bm{\beta}^\star=\arg \min_{\bm{\beta}} \hspace{0.5cm}\left \|^B\textbf{\textit{f}}^E(\bm{\beta}) - ^B\textbf{\textit{r}}^E \right\|^2_2\\
    \label{IK1}
    &\hspace{0.8cm}\text{subject to    }  \bm{\beta}_{min}\leq \bm{\beta} \leq \bm{\beta}_{max}\\
    \label{IK2}
    & \hspace{2.3cm}\Dot{\bm{\beta}}_{min}\leq \Dot{\bm{\beta}} \leq \Dot{\bm{\beta}}_{max}
\end{align}

The above optimization problem is non convex in nature. Sequential quadratic programming \cite{boggs1995sequential} approach can be used to find the solution of this nonlinear optimization problem. Finding a good solution of this optimization problem is very hard as it may often get stuck at the local optima and therefore can fail to find a feasible solution of the optimization problem if it exists. But this approach gives more freedom to handle the joint limits, joint velocity limits (defined in Eq.(3)-Eq.(4)) and other conditions by adding constraints in the optimization problem. In our work we exploited this characteristic.

\section{Collaborative Object Transportation}
\label{sec3}
\subsection{Trajectory Tracking}

Trajectory Tracking for each of the robots is first carried out independently as presented here.
In this stage, it is considered that the robots ($R_1$ and $R_2$) do not have any information about each other's position and the collaborative object transportation problem is divided into two separate planning problems for each robot.

Numerous approaches have been tried out over the years to attain trajectory tracking of mobile manipulators. We focus our attention on one such approach, which has proven to be simple, yet effective. 
The Proportional Navigation Guidance Law\cite{murtaugh1966fundamentals}; widely used in homing missile applications, is used to predict the possibility of collision of two objects based on the direct Line-of-Sight between them. When the velocity vector of the pursuer changes at the rate of rotation of the target, a collision is predicted.

Given that each of the mobile bases starts at their first respective projected desired point, the subsequent desired point acts as a stationary target for the mobile base at that particular time-step. Accordingly, the requisite Line of Sight (LOS) is calculated at each time-step to update the angular velocity $\omega$ of the base, while keeping the forward velocity, $v$ constant. As the mobile base finally enters within the radius of the aforementioned reachability distance $\rho_d$ of the current target point, the target switches to the next projected desired point on the trajectory. This iterative procedure is carried out until the robots reach their final destination. 

The rate of change of LOS is taken as $\dot{\lambda}$. With respect to the Proportional Guidance Law, we may then write:

 $\alpha_n = N\dot{\lambda}v$
; where $\alpha_n$ is the generated acceleration perpendicular to the instantaneous velocity vector of the mobile base, and N is a proportionality constant, taken as 6 in our case.

Assuming that the robots cover a circular arc trajectory at each time-step (given $a_n \neq 0$), the generated acceleration acts with acentripetal nature on the mobile base. Accordingly, the value of $\omega$ at that timestep can be generated as:

$\omega = \alpha_n/v$

Having thus obtained the values: [v, $\omega$] for the current timestep, the forward kinematics routine for the non-holonomic base, as explained before, is applied to obtain the next position $\textbf{b}^a_k$ for each robot.

\subsection{Collaborative Load Transportation}

Assuming that both the desired trajectories have same number of points $n_d$, the mobile base has to traverse repeatedly at each time step, until it enters the projected circle of reachability distance $\rho_d$ of a particular target point, as defined previously.

 This approach may cause each robot to generate different number of base positions: $\textbf{b}^a$ between their respective subsequent projected points within each desired base trajectory-- especially if the distances between the respective target points are significantly dissimilar in the defined base trajectories.
Assuming similar velocity limits in both robots, this causes a loss of correspondence between the robots-- thus violating the load rigidity constraint in our assumptions, given that the bases move so far apart as to hinder manipulation by the manipulator arms. To account for this discrepancy, we adopt a \emph{“Stop-and-Sync”} Algorithm to morph the generated velocity and base locations at each time step. This algorithm is outlined in further detail below.

\subsubsection{Stop-and-Sync Algorithm}
For each base position obtained by the robots at a particular time step, the nearest point on the desired end-effector trajectory curve is obtained through a projections of $\textbf{b}^a_k$ on $T_a$. Here, the desired end-effector trajectory $T_a$ is approximated into a set of lines through linear interpolation between the desired points. Henceforth, perpendicular projections are drawn from the current base position to the aforementioned interpolated curve. The computed intersection point between the projected perpendicular, and the interpolated line then acts as the candidate point for solving the inverse kinematics of the manipulator arm with respect to the current base position; at the current time step. 

When a robot base enters within the defined reachability radius, the corresponding target point $p$ shifts to the next point $p+1$ on the desired trajectory. At this time-step, the base location of the other robot is checked. If the target point of the other robot has not changed in a similar fashion yet, the former robot base is stopped; until the target point of the other robot changes – at which point of time, the two robots start moving together again towards their respective new target points. This approach necessitates that the two robots start moving synchronously after target switching occurs from the point $p$ to $p+1$ on both trajectories $P_a$. The Stop-and-Sync algorithm is summarized in Algorithm - \ref{alg1}.  
\begin{algorithm}
\caption{Stop-and-Sync Algorithm}\label{alg1}
\begin{algorithmic}[1]
    \State \textbf{Input}: \newline
    $b^1_0, b^2_0$: initial base positions of robots \newline
    $d^1, d^2$: discretized points of desired base trajectories $B_1$ and $B_2$ 
    
    \State \textbf{Output}: \newline
    $V_1, V_2$: generated velocities of the robots \newline
    $b^1, b^2$: generated base positions of robots \newline

    \State $R_{tol} = \gamma*\rho_l$
    \State $stop_1 = 0$ \quad \quad $stop_2 = 0$
    \State $n_{d.p} = size(d_1)$
    \While {$p \neq n_{d.p} +1$}
    \For{each timestep $k$}
    \For{each agent $a$}
    \If {$stop_a \neq 1$}
    \State Calculate velocity: $V^a_k = [v^*, \omega^*]$, and the base position: $b^a_{k+1}$
    \Else
    \State $V^a_k = [0,0]$
    \State $b^a_{k+1}$ = $b^a_k$
    \EndIf
    \State $dist_a = \lVert {d_p^a}- b^a_{k+1} \rVert$
    \EndFor
    \If {$dist_1 \leq R_{tol} \And dist_2 \leq R_{tol}$}
    \State $p = p+1$
    \State $stop_1 = 0$
    \State $stop_2 = 0$
    \ElsIf {$dist_1 \leq R_{tol} \And dist_2 > R_{tol}$}
    \State $stop_1 = 1$
    \ElsIf {$dist_1 > R_{tol} \And dist_2 \leq R_{tol}$}
    \State $stop_2 = 1$
    \EndIf
    \EndFor
    \EndWhile
\end{algorithmic}
\end{algorithm}

\subsubsection{Leader Follower Approach}

In collaborative object transportation with rigid load, the load rigidity constraint should be given more importance than the tracking error. We propose a leader follower approach that uses the information of the end effector position of one of the mobile manipulator (leader) to solve the inverse kinematics of the other manipulator arm (follower) to satisfy the load length condition, i.e. the Euclidean distance between the estimated end effector position of the leader and the end effector of the follower with respect to the world frame should be equal to the load length, $l$.       
\begin{equation}
    \label{load}
    \left\|\hat{\textbf{\textit{r}}}^{E_L} - \textbf{\textit{f}}\left(\bm{\beta}_F\right) \right\|_2=l
\end{equation}
When the two mobile manipulators reach inside the reachability distance, the leader robot, $R_L$ computes the joint angles by solving the inverse kinematics problem define in Eq. (\ref{IK})-Eq. (\ref{IK2}). Further, the end effector position of the leader, $\hat{\textbf{\textit{r}}}^{E_L}$ can also be estimated. Then the joint angles of the follower can be determined by solving the following constraint optimization problem,  
\begin{align}
    \label{ik3}
    &\bm{\beta}^*_F =\hspace{0.2cm}\arg \min_{\bm{\beta}_F}\hspace{0.4cm}\left\|^{B_F} \textbf{\textit{f}}^{E_F}\left(\bm{\beta}_F\right)-^{B_F}\textbf{\textit{r}}^{E_F}\right\|^2_2\\
    \label{ik4}
    &\hspace{0.9cm}\text{subject to}\hspace{0.3cm} \left|\hspace{0.1cm}\left\|\hat{\textbf{\textit{r}}}^{E_L} - \textbf{\textit{f}}\left(\bm{\beta}_F\right)\right\|_2 -l\right|=0\\
    \label{ik5}
    &\hspace{2.6cm}\bm{\beta}_{min}\leq \bm{\beta} \leq \bm{\beta}_{max}\\
    \label{ik6}
    & \hspace{2.6cm}\Dot{\bm{\beta}}_{min}\leq \Dot{\bm{\beta}} \leq \Dot{\bm{\beta}}_{max}
\end{align}
The leader follower algorithm is summarized in Algorithm -\ref{alg2}.
The schematic diagram of the proposed approach is shown in fig. \ref{lf}. 
    
\begin{algorithm}
\caption{Leader Follower Approach}\label{alg2}
\begin{algorithmic}[1]
    \State \textbf{Input}: $\bm{\beta}_{min}$, $\bm{\beta}_{max}$, $\Dot{\bm{\beta}}_{min}$, $\Dot{\bm{\beta}}_{max}$, $^{B_L}\textbf{\textit{r}}^{E_L}$, $^{B_F}\textbf{\textit{r}}^{E_F}$, $l$.
    \State \textbf{Output}: $\bm{\beta}^\star_L$,$\bm{\beta}^\star_F$
    \State At $k+1$, $\forall k=0,1,\ldots, N-1$;
    \State Leader: Determine $\bm{\beta}^*_L$ solving $(\ref{IK})-(\ref{IK2})$;
    \State Leader: Estimate $^{B_L}\hat{\textbf{\textit{r}}}^{E_L}$; 
    \State Follower: Determine $\bm{\beta}^\star_F$ from $(\ref{ik3})-(\ref{ik6})$.
    
\end{algorithmic}
\end{algorithm}

\begin{figure}[t!]
    \centering
    \includegraphics[width=1.\linewidth, height=0.6\linewidth]{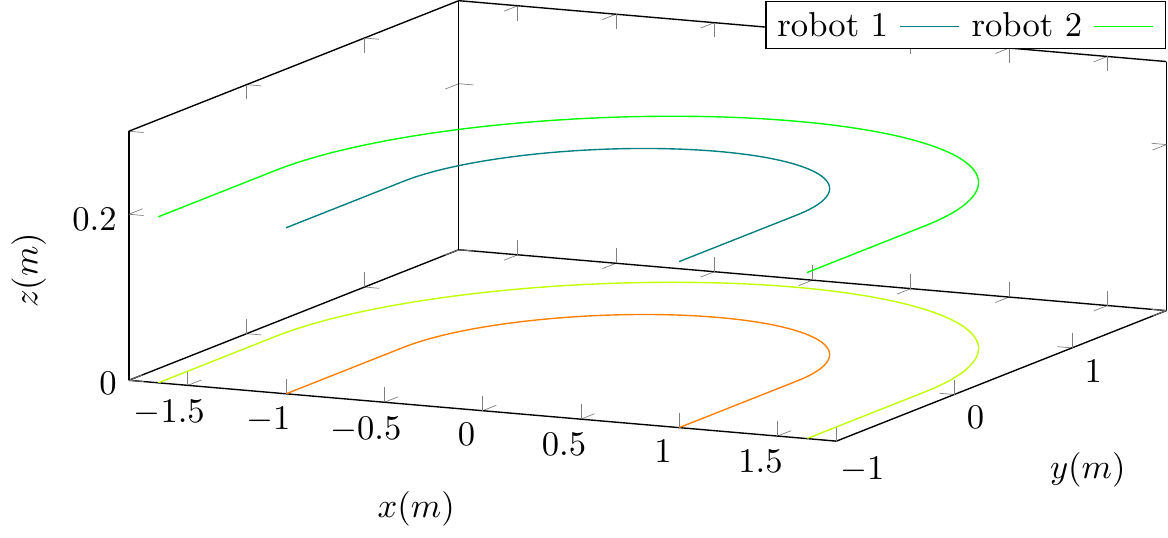}
    \caption{The desired trajectories and their projection on the $X-Y$ plane of the two robots.  }
    \label{traj}
\end{figure}
\section{Simulation and Results}
\label{sec4}

\subsection{Simulation Setup}
The proposed algorithm is tested on a simulated environment in MATLAB to capture their efficiency. For the mobile manipulator system, we use the specifications for the TurtleBot3 Waffle Pi with the OpenMANIPULATOR-X attached atop it. The pair of robots is assumed to carry a slender load of length 0.65 m. The control variables $[v, \omega]$ are passed on to the robots with a time gap, $\Delta t$ of $0.08$ seconds. For the manipulator arm we incorporate the following constraints on joint angles $(\bm{\beta})$ in rad and rate of change of the joint angles $\dot{\bm{\beta}}$ in rad/sec.:  $[-0.9\pi,-0.57\pi,-0.3\pi,-0.5\pi]^\intercal \leq \bm{\beta}^\intercal \leq [0.9\pi,0.5\pi,0.44\pi,0.65\pi]^\intercal$ and $\dot{\bm{\beta}}_i=4 \hspace{0.3cm}\forall i \in \{ 1,2,3,4\}$. We further limit the gripper orientation of the manipulator arm such that it always remains parallel to the $X-Y$ plane while carrying the load for hardware limitation.   

Since most generated trajectories can be broken down into a straight line and circular arcs, we have, for generality, considered the desired trajectories of the two robots as shown in fig. \ref{traj} for simulation. The desired trajectory of each robot has three parts, a semicircle in the middle and two straight lines of length 1m at the both end of the semicircle. The radius of the semicircle for inner and outer robot is taken 1 m and 1.65 m respectively. The height along the z axis of each trajectory is taken as 0.2m. The initial positions of the base of the mobile manipulators are $ (1,-1)$ and $(1.65,-1)$. The goal positions of the them are $(-1,-1)$ and $(-1.65,-1)$. The reachability tolerance value $\gamma$ is taken as 0.4. The time delay in the system between computing the solution of inverse kinematics of the leader and the follower is considered negligible.    
\begin{figure*}[ht]
    \begin{subfigure}[b]{0.32\linewidth}
         \centering
         \includegraphics[height=0.7\linewidth,width=\linewidth]{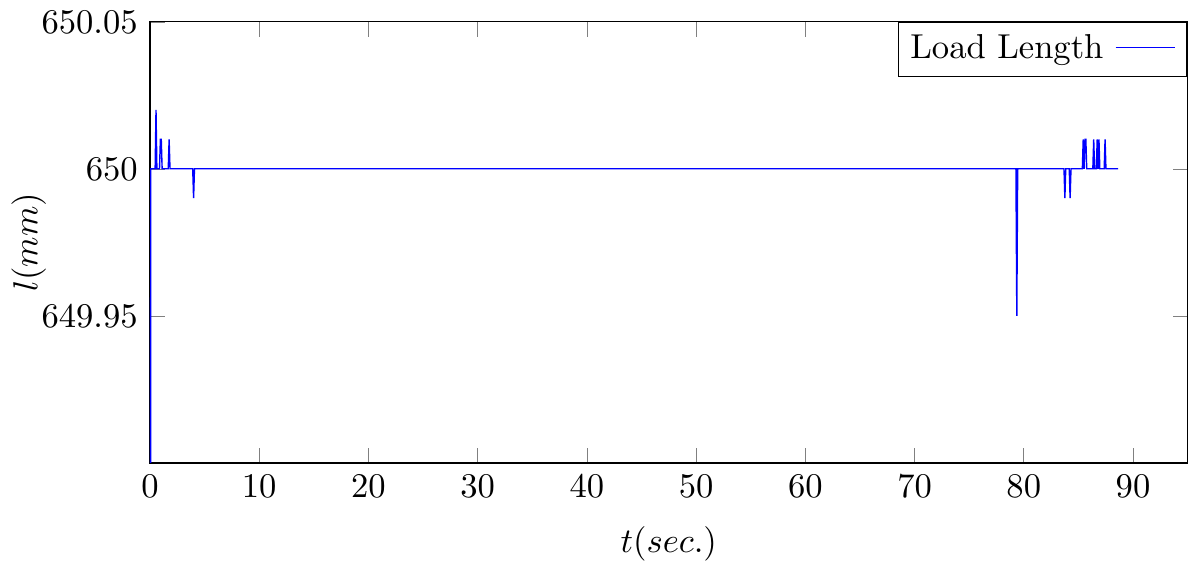}
         \caption{}
         \label{l1}
     \end{subfigure}
     \hfill
     \begin{subfigure}[b]{0.32\linewidth}
         \centering
         \includegraphics[height=0.7\linewidth,width=\linewidth]{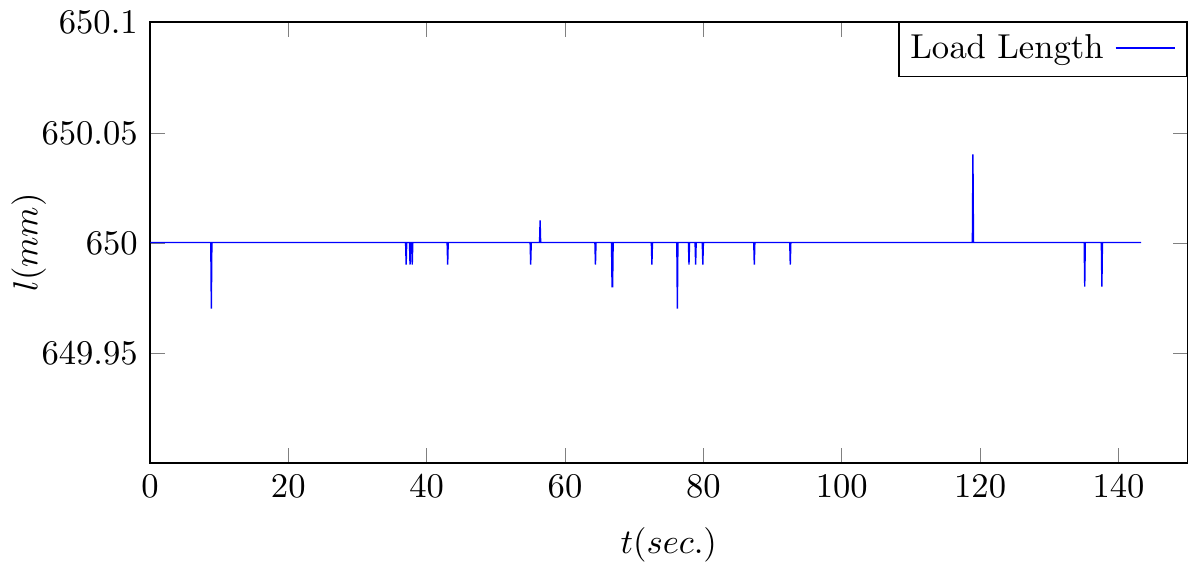}
         \caption{}
         \label{l2}
     \end{subfigure}
     \hfill
     \begin{subfigure}[b]{0.32\linewidth}
         \centering
         \includegraphics[height=0.7\linewidth,width=1\linewidth]{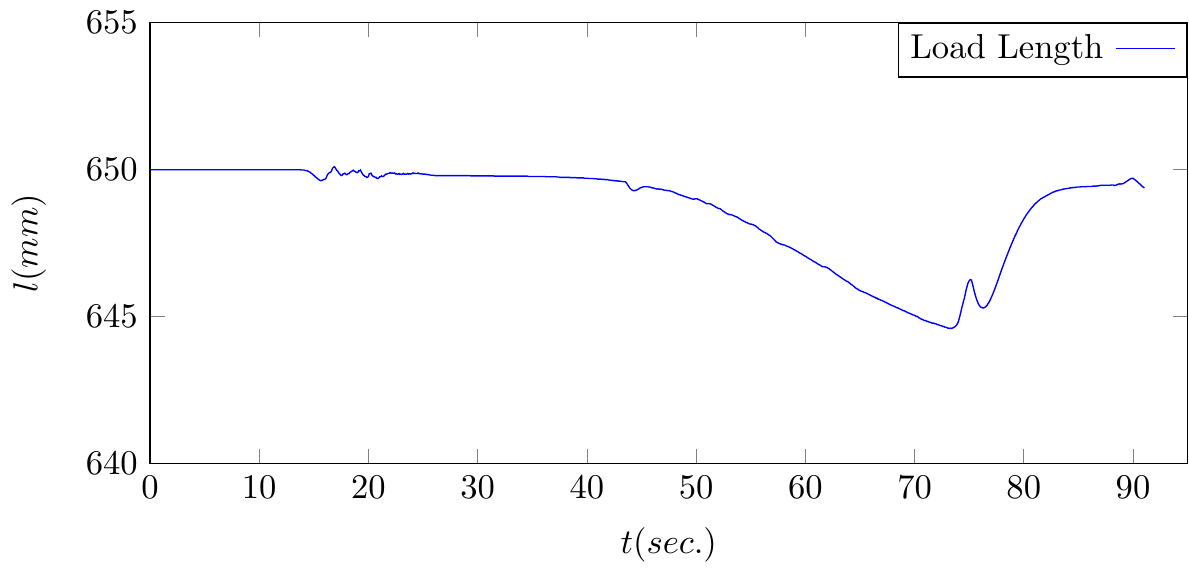}
         \caption{}
         \label{l3}
     \end{subfigure}
   \caption{\small The obtained distances between the end-effectors of the two robots for the three different methods. Fig. \ref{l1}, fig. \ref{l2} and fig.\ref{l3} represent the results from PN Guidance, RRT generation and MPC based methods respectively.}
   \label{l}
\end{figure*}

\begin{figure*}[ht]
    \begin{subfigure}[b]{0.32\linewidth}
         \centering
         \includegraphics[height=0.7\linewidth,width=\linewidth]{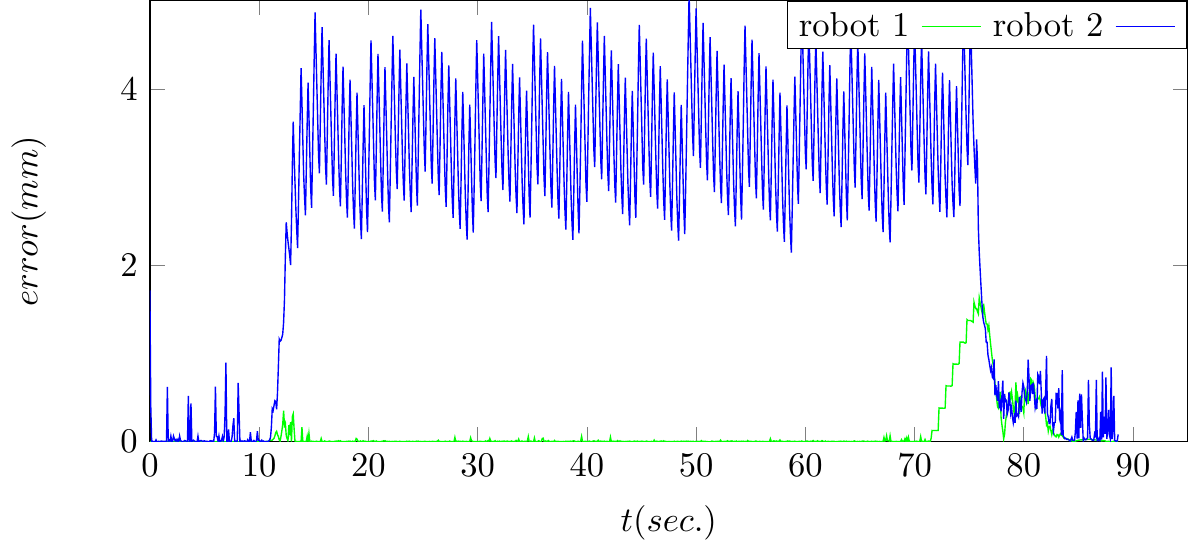}
         \caption{}
         \label{m1}
     \end{subfigure}
     \hfill
     \begin{subfigure}[b]{0.32\linewidth}
         \centering
         \includegraphics[height=0.7\linewidth,width=\linewidth]{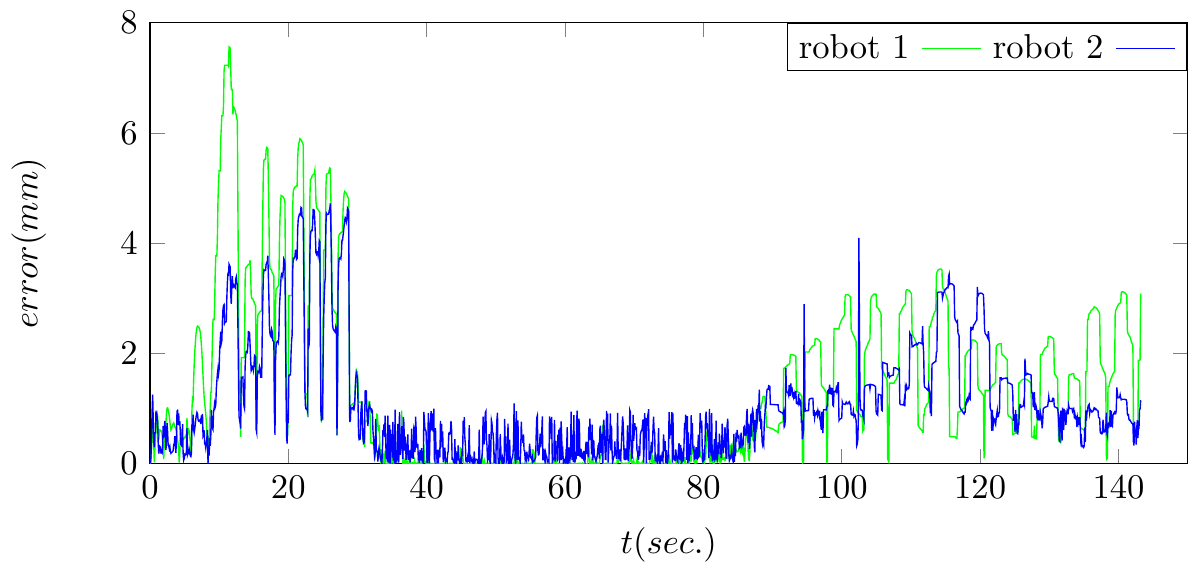}
         \caption{}
         \label{m2}
     \end{subfigure}
     \hfill
     \begin{subfigure}[b]{0.32\linewidth}
         \centering
         \includegraphics[height=0.7\linewidth,width=1\linewidth]{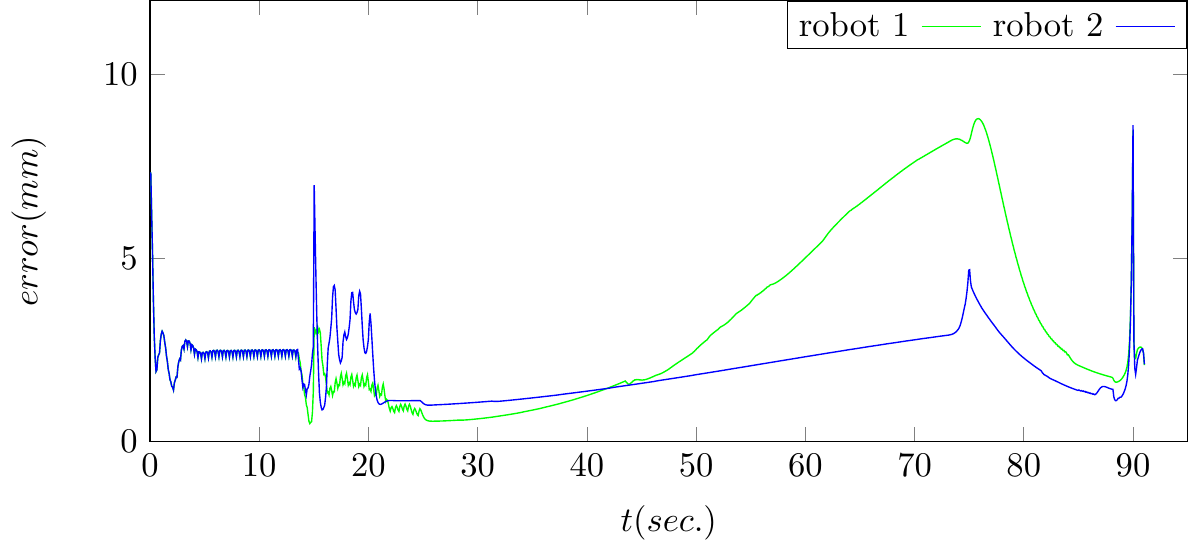}
         \caption{}
         \label{m3}
     \end{subfigure}
   \caption{\small  The trajectory tracking errors of two robots for the three methods.  Fig. \ref{m1}, fig. \ref{m2} and fig. \ref{m3}
represents the result from PN Guidance, RRT generation and MPC based methods respectively }
\label{m}
\end{figure*}

\begin{figure*}[ht]
    \begin{subfigure}[b]{0.32\textwidth}
         \centering
         \includegraphics[height=0.7 \textwidth,width=\textwidth]{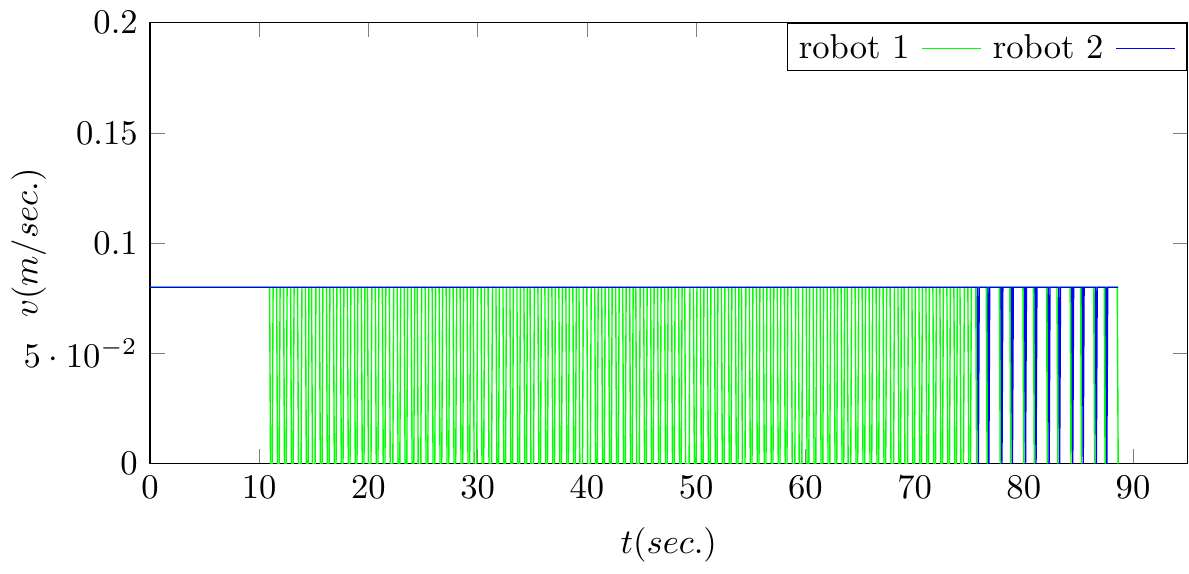}
         \caption{}
         \label{v1}
     \end{subfigure}
     \hfill
     \begin{subfigure}[b]{0.32\textwidth}
         \centering
         \includegraphics[height=0.7 \textwidth,width=\textwidth]{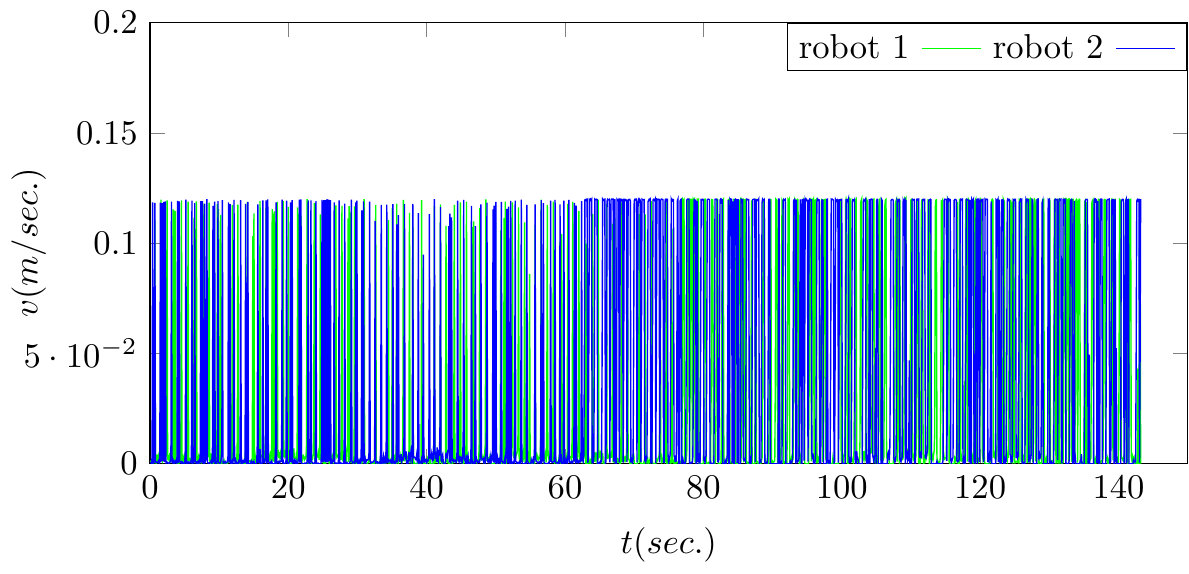}
         \caption{}
         \label{v2}
     \end{subfigure}
     \hfill
     \begin{subfigure}[b]{0.32\textwidth}
         \centering
         \includegraphics[height=0.7 \textwidth,width=\textwidth]{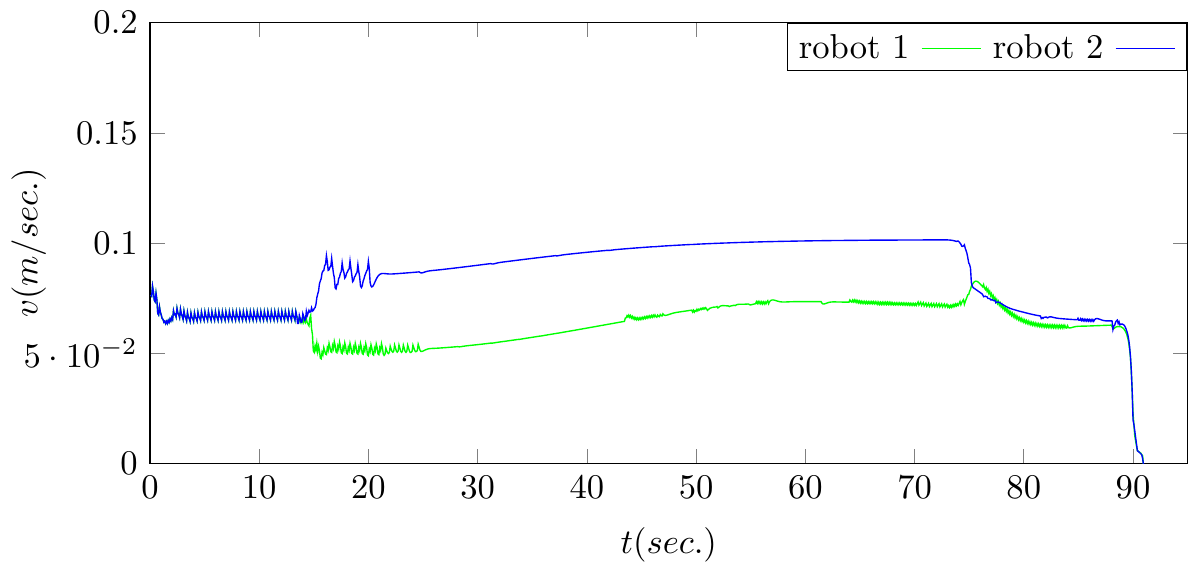}
         \caption{}
         \label{v3}
     \end{subfigure}
   \caption{\small  The linear velocities of the two robots for the three different methods. Fig. \ref{v1}, fig. \ref{v2} and fig.\ref{v3} represent the results from PN Guidance, RRT generation and MPC based methods respectively. }
\label{v}
\end{figure*}

\begin{figure*}[ht]
    \begin{subfigure}[b]{0.32\textwidth}
         \centering
         \includegraphics[height=0.7 \textwidth,width=\textwidth]{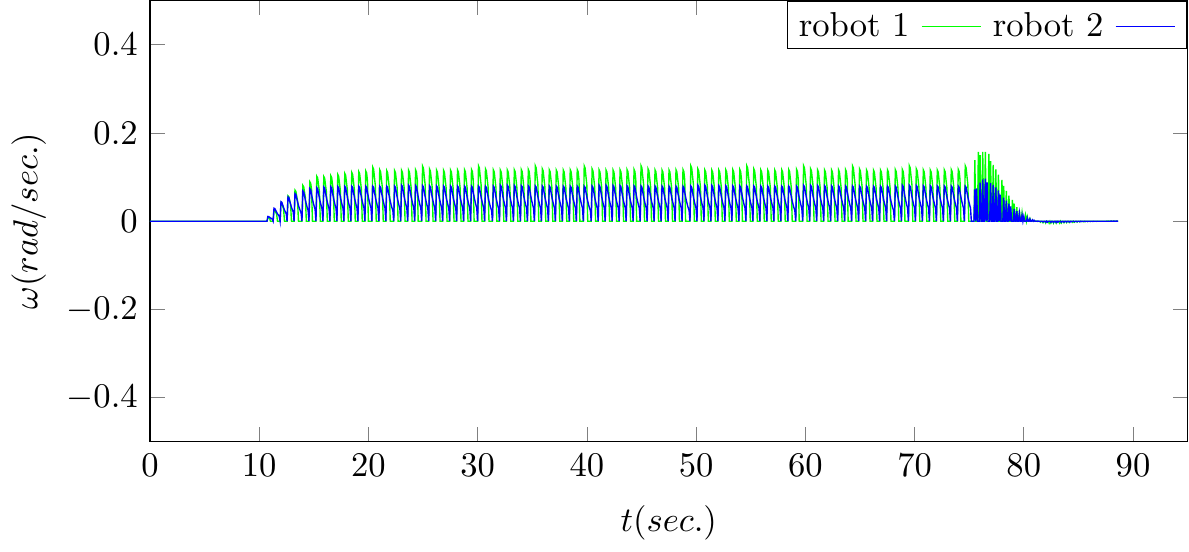}
         \caption{}
         \label{w1}
     \end{subfigure}
     \hfill
     \begin{subfigure}[b]{0.32\textwidth}
         \centering
         \includegraphics[height=0.7 \textwidth,width=\textwidth]{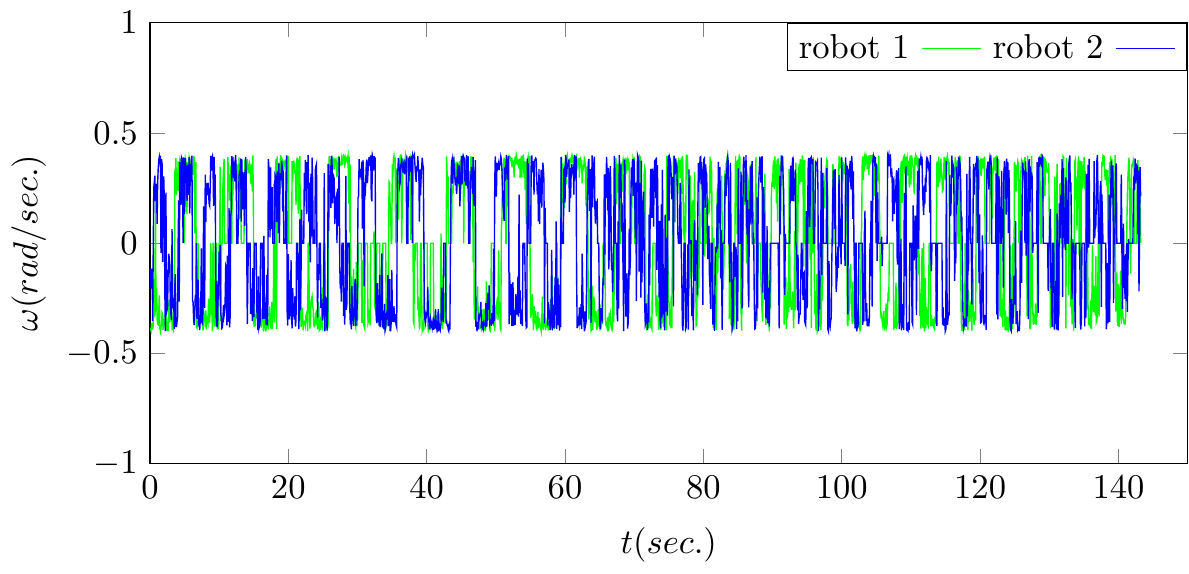}
         \caption{}
         \label{w2}
     \end{subfigure}
     \hfill
     \begin{subfigure}[b]{0.32\textwidth}
         \centering
         \includegraphics[height=0.7 \textwidth,width=\textwidth]{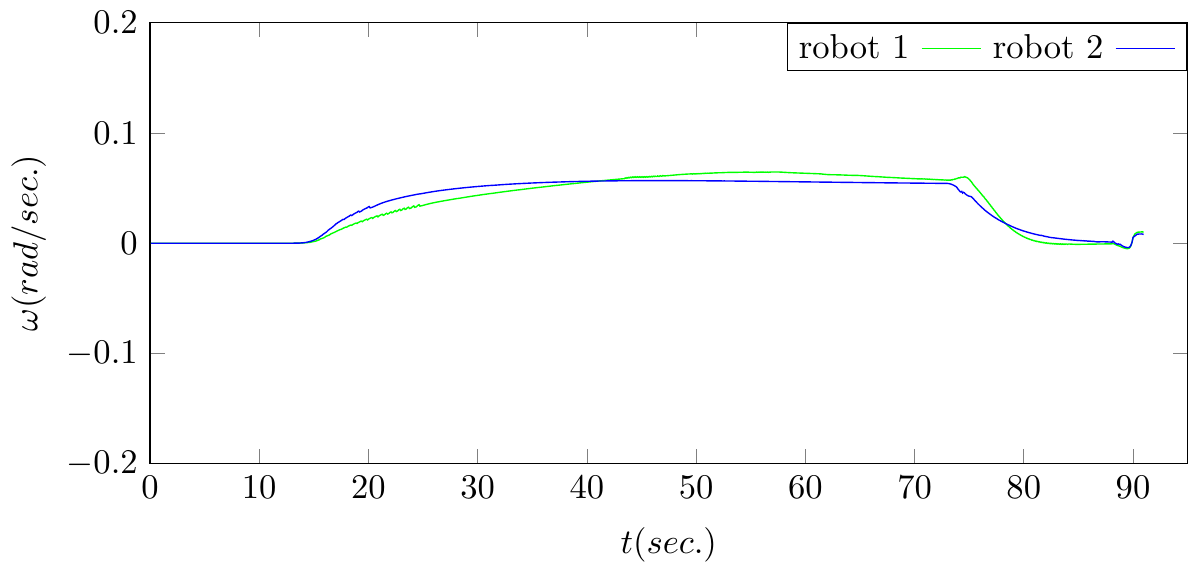}
         \caption{}
         \label{w3}
     \end{subfigure}
   \caption{\small The angular velocities of the two robots for the three different methods. Fig. \ref{w1}, fig. \ref{w2} and fig.\ref{w3} represent the results from PN Guidance, RRT generation and MPC based methods respectively. }
\label{w}
\end{figure*}

\begin{figure*}[t]
     \centering
     \begin{subfigure}[b]{1\textwidth}
         \centering
         \includegraphics[width=\textwidth]{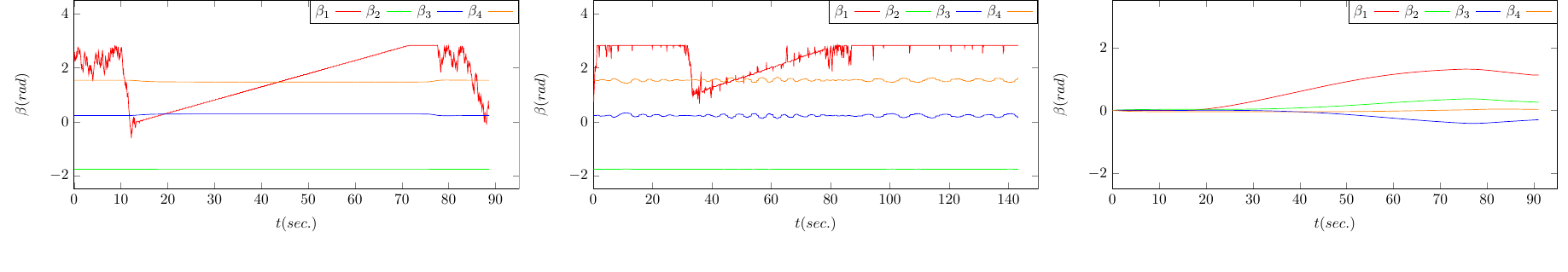}
         \label{b1}
     \end{subfigure}
     \hfill
     \begin{subfigure}[b]{1\textwidth}
         \centering
         \includegraphics[width=\textwidth]{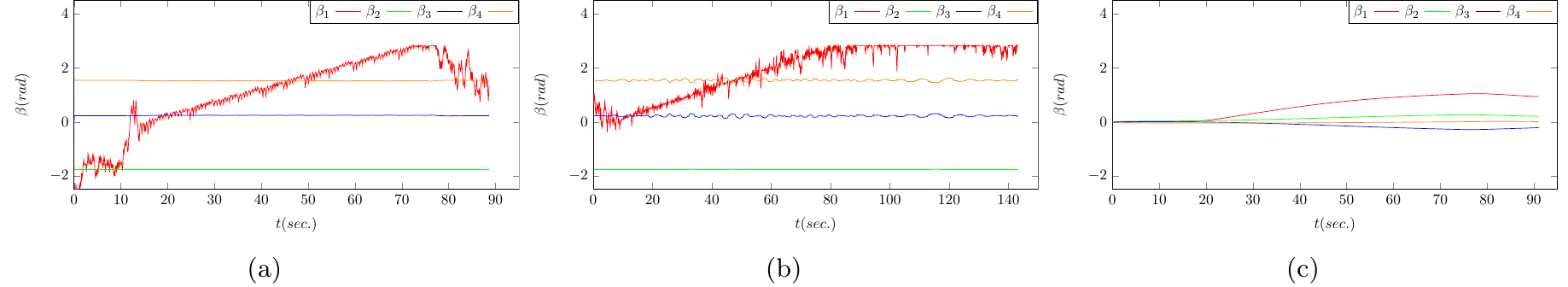}
         \label{b2}
     \end{subfigure}
        \caption{\small The joint angles of the two robots for the three methods. The top plot portrays the joint angles of robot 1 while the bottom plot displays the same for robot 2. Fig.9a, fig.9b and fig.9c represent the results for PNG based, RRT generation based and MPC based approaches respectively.}
        \label{b}
\end{figure*}

\subsection{Comparative Approaches}
The presented approach is brought into comparison with two other approaches, to compare efficacy of the results with respect to the ascertained parameters. 

In the first approach, an attempt is made to generate a single Rapidly-exploring random tree (RRT) from the given start node to the final goal node, with the intermediate discretized points $d^a_P$ acting as the transitional target nodes at each timestep. 
This process is carried out iteratively until the tree spans the entire trajectory $T^a$.    

Given a point $(x^d_{p},y^d_{p})$ on the desired trajectory, each robot first generates 500 random sets of linear and angular velocities: $[v, \omega]$ within a specified range. Each velocity is registered upon the current base location of each robot, and the corresponding final position is generated upon the application of the velocities for time $\Delta t$, all the while respecting the non-holonomic constraints of the robots. Among the generated subsequent locations, the one nearest to the aforementioned desired point $d^a_P$; and the equivalent velocity data: $[v^*, \omega^*]$ is chosen for the current timestep.

The computation of the control variables for the agent $a$, can thus, be formulated as an optimization problem that minimizes the following objective function:
\begin{equation}
    \label{J}
    J=\left|d-\rho_d\right|
\end{equation}

where:
\begin{equation}
    \label{d}
    d=\sqrt{(x^{d,a}_{p}-x^a_k)^2+(y^{d,a}_{p}-y^a_k)^2
    }
\end{equation}

The objective function indicates the control velocities needed to reach within the reachability distance of the next desired point. The optimization problem is defined below,
\begin{equation}
    \label{v}
    v^\star,\omega^\star = \arg \min_{v,\omega} J
\end{equation}

Once path generation is achieved in this manner, prior approaches of the Stop-and-Sync Algorithm, and the Leader-Follower are incorporated to bring about collaborative load transport by the robots.

The presented approach is also compared with a Sequential Linear Quadratic Model Predictive \cite{Neunert2016} approach in simulation wherein each robot individually tracks their own reference trajectory based on a desired object trajectory. The model used is given by
\begin{equation}
  \dot{\pmb{x}} =\begin{bmatrix}
      v\cos{\theta}\\
      \omega\sin{\theta}\\
      \dot{\pmb{\beta}}
  \end{bmatrix}    
\label{mm_kinematics}
\end{equation}
where the state $\pmb{x}\triangleq[x,y,\theta,\pmb{\beta}^T]^T \in \mathbb{R}^{3+n}$ and  $\pmb{u} \triangleq [v,\omega,\dot{\pmb{\beta}^T}]^T\in \mathbb{R}^{2+n}$ are the linear and angular velocities for the base and joint velocities respectively.
For each robot, the cost function considered over a horizon of length $N$ is 
\begin{equation}
    \label{moc_cost}
    J = \pmb{ee}(\pmb{x}_N)^{T}\pmb{Q}\pmb{ee}(\pmb{x}_N)+\sum_{t=1}^{N-1}\pmb{ee}(\pmb{x}_t)^{T}\pmb{Q}\pmb{ee}(\pmb{x}_t) +\pmb{u}_t^{T}\pmb{R}\pmb{u}_t
\end{equation}
where the function $\pmb{ee}(\pmb{x}_t) = \pmb{FK}(\pmb{x}_t) - \pmb{r}(t)$ is the tracking error between the end effector position $\pmb{FK}(\pmb{x}_t)\in \mathbb{R}^3$ and reference trajectory $\pmb{r}(t)$ at time $t$. The matrices $\pmb{Q}$ and $\pmb{R}$ are positive diagonal weighting matrices. Input and joint angle constraints are incorporated in a \textit{soft} manner by adding relaxed barrier functions \cite{Feller2017} to the cost equation. 
The parameters of all the stated approaches are kept similar to the extent possible, for proper comparison.

\subsection{Simulation Results}
The simulation results are depicted in fig. \ref{l} -\ref{b}. The time taken to reach the goal state from the initial position differs for the three approaches. Fig. \ref{l} represents the distance between the end effectors of the mobile manipulators$(l)$. This indicates the extension and compression the rigid load may observe while the mobile manipulators are in action. The proposed leader follower with stop and wait approach ensures minimum deviation of $l$ from the desired load length $0.65$m which can also be seen in fig. \ref{l1} and fig. \ref{l2}. For SLQ MPC approach distortion of $l$ can be observed in fig. \ref{l3}.

Trajectory tracking errors for the three methods are depicted in Fig. \ref{m}. In our work, the Leader follows the trajectory $T_L$ independently, while the Follower tries to remain close to the trajectory $T_F$, while giving greater priority to maintaining a constant $l$. It is expected to observe tracking error for the Follower as seen in fig. \ref{m1} though the tracking error for the Leader is minimal. For RRT generation based approach it can be seen that the tracking error of Leader is comparable to that of follower in fig. \ref{m2} though both of them return better tracking error than the SLQ MPC approach. 

The control input to mobile base $(v,\omega)$ is shown in fig. \ref{v}-fig.\ref{w}. Due to the Stop-and-Sync algorithm fluctuations can be noticed both in $v$ and $\omega$ for PN Guidance and RRT generation based approach while for MPC based approach it appears to be smooth. For the PN Guidance based approach, the control variable $v$ remains constant for Robot-2 (which has to traverse a longer distance for the outer semicircular path), while that of Robot-1 is interrupted at regular intervals to accommodate synchronization with Robot-2. Due to difference in orientations at the end of the circular arc trajectory, both robots undergo seemingly unexpected intervals of variations in $v$ during the second straight line phase. As Robot-1 has to navigate a tighter curve with smaller radius, its values of $\omega$ are consistently higher. For RRT generation based approach the variations in $v$ due to \emph{Stop-and-Sync} can be spotted from the very beginning in both robots, due to randomizations in velocity generation. The evaluation of the joint angles are shown in fig. \ref{b} for the three methods.                                         
\subsection{Hardware Experimentations}
The experiments run on the TurtleBot 3 Waffle Pi hardware can be accessed at this \href{https://www.youtube.com/playlist?list=PLV-rN3LPVdPrkdPvWdGozuTC-oIo7ywZu}{playlist}. Here, two robots are made to work collaboratively to transport a rigid rod while following specified trajectories. 
Due to inherent hardware constraints, a horizontal constraint is imposed upon the end-effectors to prevent undue twisting of the load. As can be seen from the videos, the robots were able to carry out load transportation successfully on an eclectic variety of trajectories.

\section{CONCLUSIONS}
\label{sec5}
 In this paper, we have considered two robots collaboratively transporting a rigid load, while subject to non-holonomic constraints. To that end, a Leader-Follower Approach for the inverse kinematics solution; coupled with a Stop-and-Sync Algorithm to keep the mobile bases in close proximity has been proposed. The proposed approach, as can be ascertained from the results, shows better results with respect to our salient metrics of load length, and trajectory tracking error. Hardware experimentations also prove the efficacy of the simulation results.

 The proposed approach, due to its inherent nature, provides discontinuous profiles of velocities. In future works, it shall be attempted to smoothen out these profiles through a predictive localization technique. Interchangeability of the agents acting out as Leader and Follower shall also be attempted to obtain better results during trajectory tracking of tight curves.

\addtolength{\textheight}{-12cm}   





\bibliography{bibliography.bib}
\bibliographystyle{IEEEtran}

\end{document}